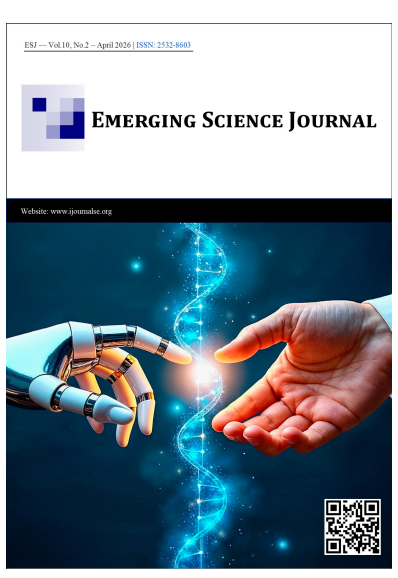

# Pattern Recognition Tasks with Personalized Federated Learning

Md. Arifur Rahman [1], Isha Das [2], Mushfiqur Rahman Abir [3], B. M. Taslimul Haque [4], Abdullah Al Noman [5], Abir Ahmed [6], Md. Jakir Hossen [7]*

[1] *College of Graduate and Professional Studies, Trine University, Angola, IN 46703, United States.*

[2] *Network Communication and IoT Lab, Chittagong University of Engineering and Technology, Chittagong, Bangladesh.*

[3] *Department of Computer Science and Engineering, American International University-Bangladesh, Dhaka, Bangladesh.*

[4] *Information Systems, Central Michigan University, New Castle, DE 19720, United States.*

[5] *Wilmington University, Alexandria, VA 22314, United States.*

[6] *Department of Information Technology, Washington University of Science & Technology, VA, United States.*

[7] *Center for Advanced Analytics (CAA), COE for Artificial Intelligence, Faculty of Engineering & Technology (FET), Multimedia University, Melaka 75450, Malaysia.*

**Abstract**

Personalized Federated Learning (PFL) constitutes a novel paradigm that tailors Machine Learning (ML) models to individual clients, thereby furnishing personalized model updates whilst upholding stringent data privacy principles. Diverging from conventional standard Federated Learning (FL) approaches, PFL adapts models to distinct client data distributions, engendering heightened levels of accuracy, customization, and data security, all while minimizing communication overhead. This methodology proves particularly salient in contexts marked by pattern recognition tasks reliant upon heterogeneous data sources and underpinned by paramount privacy apprehensions. In the present research endeavor, this article undertake a comprehensive comparative analysis of seven distinct PFL algorithms deployed across three diverse datasets, namely MNIST, SignMNIST, and Digit5. The overarching objective entails ascertaining the preeminent PFL algorithm, within the framework of pattern recognition tasks, through a rigorous evaluation anchored in metrics encompassing Accuracy, Precision, Recall, and F1 Score. Concurrently, an in-depth scrutiny of these PFL algorithms is conducted, elucidating their operative workflows, advantages, and limitations. Through empirical investigation, the findings evince that APPLE, FedGC, and FedProto emerge as stalwart contenders, consistently furnishing superior performance across the spectrum of assessed datasets, while acknowledging the contextual specificity of alternative algorithms and the potential for iterative refinement to realize optimal outcomes.

## 1- Introduction

In recent years, ML has emerged as a pivotal technology, transforming numerous industries and enabling data-driven decision-making. Despite its widespread success, traditional ML approaches often grapple with significant challenges related to data privacy, storage, and computational costs [1, 2]. Centralized learning systems rely on aggregating raw data from diverse sources into a central repository for training models, raising critical concerns about the privacy and security of sensitive data. These systems become potential targets for breaches, exposing individuals and organizations to significant risks. Moreover, centralized architectures require substantial storage capacities and robust computational infrastructure, leading to increased operational costs and challenges in managing vast datasets. The need to transfer raw data for training exacerbates bandwidth consumption and makes these systems less viable in environments with limited or unreliable network connectivity. Furthermore, compliance with stringent privacy regulations, such as GDPR and HIPAA, often poses additional hurdles for the adoption of centralized ML systems [3, 4].

* **CONTACT**: jakir.hossen@mmu.edu.my

FL has emerged as a transformative paradigm, addressing the fundamental limitations of traditional ML by decentralizing the training process and prioritizing privacy [5, 6]. Unlike centralized systems, FL enables local computations on user devices, transferring only model updates rather than raw data to a central server. This approach minimizes data transfer, reduces dependency on centralized storage, and ensures compliance with privacy regulations. By leveraging the diverse data distributed across devices, FL promotes scalability, robustness, and adaptability in a wide range of applications. However, the classical FL framework predominantly focuses on aggregating model updates to construct a global model. This global aggregation often neglects the unique preferences and data heterogeneity across users and devices, resulting in suboptimal performance and limited personalization [7].

PFL addresses these limitations by tailoring global models to individual user needs. Through user-specific model updates, PFL considers device heterogeneity, offers fine-grained privacy control, and enhances user engagement and trust in the learning process [8, 9]. By aligning model outputs with individual user preferences, PFL achieves higher accuracy and relevance, paving the way for widespread adoption in domains where personalization is critical. Applications such as healthcare diagnostics, personalized finance, and intelligent consumer devices exemplify areas where PFL's capacity to balance privacy and personalization is transforming the ML landscape. The evolution of FL has been marked by significant advancements, transitioning from a conceptual framework addressing centralized data limitations to a sophisticated decentralized system with diverse capabilities. Early algorithms, such as FedAvg [5], focused on efficient aggregation of model updates, while newer approaches, like SCAFFOLD [10] and FedProx [11], address communication efficiency and data heterogeneity. Recent innovations, including knowledge distillation [12] and PFL, have further enriched FL's potential, establishing it as a cornerstone for collaborative ML in distributed environments. In particular, FL has demonstrated exceptional promise in pattern recognition tasks, where its decentralized approach enables the development of robust and accurate models while safeguarding data privacy.

Pattern recognition, the automated identification of patterns and regularities in data, is a foundational element of numerous applications, including statistical data analysis, signal processing, image analysis, information retrieval [13], bioinformatics, data compression, and computer graphics [14]. Traditional ML methods, such as support vector machines (SVM), k-nearest neighbors (KNN), decision trees, random forests, and naive Bayes, have been extensively applied to pattern recognition tasks [15, 16]. However, these methods are inherently centralized, raising significant concerns about data privacy and security, particularly when handling sensitive information.

FL has redefined the landscape of pattern recognition by enabling decentralized model training on user devices, thus ensuring data privacy. By employing federated optimization, FL consolidates model updates while keeping personal data localized, enabling collaborative development of recognition models across domains like healthcare, smart devices, and finance. This decentralized strategy not only ensures privacy but also addresses the heterogeneity of distributed data, making FL an invaluable tool for privacy-sensitive applications [17]. Building on this foundation, PFL further extends FL's capabilities, allowing recognition models to be personalized based on user-specific preferences and requirements. This personalized approach significantly enhances accuracy and relevance, addressing the limitations of one-size-fits-all global models.

The need for privacy-aware, personalized solutions continues to grow as digital systems increasingly integrate into sensitive and regulated domains [18, 19]. In this context, research on PFL becomes highly relevant, offering a novel approach to balancing privacy, personalization, and performance. A systematic evaluation of PFL methods against traditional ML and FL frameworks is critical to understanding their effectiveness in various scenarios. Such research not only highlights the benefits of personalization in FL but also provides valuable insights for practitioners, researchers, and policymakers seeking to optimize ML in distributed, privacy-sensitive environments. Analyzing the performance of PFL algorithms in pattern recognition tasks, this study aims to fill a critical gap in the literature, advancing the understanding of personalized solutions in federated environments. This exploration will inform the development of next generation ML systems that are efficient, adaptable, and privacy-conscious, thereby meeting the evolving demands of a data-driven world.

This paper analyzed seven PFL algorithms and compared their performance on three different datasets related to pattern recognition tasks. The PFL algorithms are APPLE [20], FedALA [21], FedBABU [22], FedGC [23], FedPAC [24], FedPCL [25] and FedProto [26].

The contributions of this paper are:

- An exhaustive deliberation on seven distinct PFL algorithms, providing a comprehensive exposition of their operational methodologies, accompanied by an elucidation of their inherent advantages and disadvantages.
- Evaluation of the performance of these PFL algorithms is conducted across multiple datasets, specifically MNIST [27], SignMNIST [28], and Digit5 [29]. This rigorous analysis furnishes valuable insights into their relative efficacy when applied to pattern recognition tasks, substantiated by the utilization of pivotal performance metrics.
- Subsequently, the paper performs a comparative analysis of these PFL algorithms against established paradigms such as traditional ML and standard FL, aiming to enhance the understanding of PFL algorithm performance in pattern recognition tasks.

- Finally, it presents a trade-off between performance, security, and data privacy with potential limitations and future directions.

The rest of the paper is structured as follows: Section 2 discussed the background related to our work. Section 3 described the detailed methodology of our experiment, including dataset overviewing, experimental setup, data preprocessing, algorithms, and the workflow for doing the experiment. Then, Section 4 analyzed the results of our experiment, including an abstract comparison of our results with traditional ML and FL algorithms. Later, Section 5 discussed the trade-off between performance, security, and data privacy with potential limitations and future directions. Finally, Section 6 concludes the paper.

## 2- Background

Pattern recognition tasks have a rich history that spans several decades. Traditionally, these tasks were primarily approached through classical ML algorithms before the advent of Deep Learning (DL) and FL techniques. In the early days of pattern recognition (1960s-1990s), researchers relied on handcrafted feature extraction methods and conventional ML algorithms to solve image-related tasks. These tasks included object recognition [30], character recognition, and face detection. Feature extraction involved manually designing filters and feature descriptors to represent visual patterns in images. The extracted features were then fed into classifiers like Support Vector Machine (SVM), k-Nearest Neighbors (k-NN), and Decision Trees to make predictions. While these approaches showed promising results for specific tasks, they often lacked generalization and struggled to handle complex and diverse datasets.

The breakthrough in pattern recognition came with the rise of DL (DL) in the mid-2000s. Convolutional Neural Networks (CNNs) emerged as a dominant approach, significantly improving performance on various tasks due to their ability to automatically learn hierarchical representations from raw pixel data. This led to dramatic advancements in tasks like object pattern recognition, and semantic segmentation.

As DL models became more powerful, researchers explored ways to train them on decentralized data without centralizing sensitive information. This gave rise to FL, a distributed ML paradigm. FL enables model training across multiple devices or servers without sharing raw data. Instead, only model updates are exchanged, making it privacy-preserving. This approach has been particularly useful in scenarios like mobile devices, where data privacy is paramount. FL found applications in pattern recognition tasks as well. For example, in surveillance systems, where cameras are spread across various locations, local models on each camera could be trained using FL while preserving data privacy. Additionally, FL can be utilized in medical imaging, where hospitals can collaboratively improve a model for diagnosing diseases without sharing patient data.

Table 1 provides insights into the performance and advancements of traditional ML algorithms for pattern recognition tasks across different research works. Each row represents a scientific paper, with the columns indicating the algorithms used, the datasets employed for evaluation, the achieved results in terms of accuracy or error rates, and the core contributions of each study. The algorithms include DP-SVM, CNN, KNN, Gaussian Naïve Bayes, Decision Tree, GPFL, and Hybrid-RF, while the datasets range from MNIST to Sign-MNIST. The core contributions of the papers involve sign language gesture recognition, optical character recognition, and novel approaches like Genetic Program Feature Learner and Hybrid models.

**Table 1. Work done on different traditional ML algorithms on different pattern recognition tasks**

| References | Algorithms | Datasets | Results (%) | Core Contributions |
|---|---|---|---|---|
| [31] | DP-SVM | MNIST | Linear SVM = 97.2, RBF SVM = 98.3 | The core contribution of the scientific paper is the development of a privacy-preserving image pattern recognition scheme using Support Vector Machine (SVM) and Differential Privacy (DP) framework. |
| [32] | CNN | Sign-MNIST | Accuracy = 93 | The core contribution of the scientific paper is the development of a DL based application that uses a custom Convolutional Neural Network (CNN) to recognize sign language gestures from video frames. |
| [33] | KNN | MNIST | Sensitivity = 97.18, Precision = 97.92, Specificity = 98.58 | The core contribution of the scientific paper is the development and evaluation of a system for Optical Character Recognition (OCR) focused on handwritten digits. The paper compares the performance of ML and DL algorithms, specifically the CNN classifier, for accurately extracting and classifying handwritten letters and numbers from images. |
| [34] | Gaussian Naïve Bayes | Sign-MNIST | Accuracy = 46.85 | The core contribution of the scientific paper is a comprehensive analysis and comparative study of various pattern recognition techniques for gesture recognition. The researchers evaluated popular classifiers such as Naïve Bayes, K-Nearest Neighbor (KNN), random forest, XG-Boost, Support vector classifier (SVC), logistic regression, Stochastic Gradient Descent Classifier (SGDC), and Convolution Neural Networks (CNN). |
| [35] | Decision Tree | MNIST | True-Classified Scenario = 90.37, False-Classified Scenario = 9.63 | The core contribution of the scientific paper is the proposal of a method for offline handwritten digit recognition using ML algorithms. Various ML methods, such as Support Vector Machine, Multilayer Perceptron, Decision Tree, Naïve Bayes, K-Nearest Neighbor, and Random Forest, are employed and evaluated for this purpose. |
| [36] | GPFL | MNIST | Accuracy = 91.66 | The core contribution of the scientific paper is the introduction of Genetic Program Feature Learner (GPFL), a novel generative GP feature learner for 2D images. GPFL uses multiple GP runs to generate models that focus on specific high-level features of training images. These models are then combined into a function that reconstructs observed images. |
| [37] | Hybrid-RF | MNIST | Error = 1.97 | The core contribution of the scientific paper is the proposal of a hybrid model that combines a Convolutional Neural Network (CNN) and a Random Forest (RF) for image pattern recognition. Instead of using the gradient algorithm to adjust CNN parameters, the model inputs CNN-extracted features into RF for classification. |

Table 2 provides a valuable insight into the evolution and results of traditional FL algorithms for pattern recognition tasks across different research efforts. Each row in the table represents the academic papers, and the corresponding columns show the algorithmic choices, the datasets used for evaluation, the resulting accuracy metrics, and the key contributions of each study. The algorithms selected are FedAvg, SCAFFOLD, FedProx, FedDyn, MOON and FedGen, while the datasets evaluated are MNIST, EMNIST, FEMNIST, Shakespeare, CIFAR-10, CIFAR-100, Sent140, Tiny-Imagenet and Celeba. The core contributions of the papers involve decentralized training on distributed mobile data, focusing on minimized communication costs and robust performance, maintaining stable convergence in the face of data heterogeneity, handling data diversity and robustness, combination of model contrast and data-free distillation techniques to improve FL in different scenarios.

**Table 2. Work done of different traditional FL algorithms on different patterns**

| References | Algorithms | Datasets | Results (%) | Core Contributions |
|---|---|---|---|---|
| [5] | FedAvg | MNIST, Shakespeare, CIFAR-10 | Accuracy: MNIST=99, Shakespeare=54, CIFAR- 10=85 | The core contribution of this paper is the introduction of FL, a decentralized method for training deep networks on distributed mobile data, minimizing communication costs while ensuring strong performance across diverse data distributions. |
| [10] | SCAFFOLDE | MNIST | Similarity=84.2 | The core contribution of this paper is the SCAFFOLD algorithm, which addresses the 'client-drift' issue in FL by using control variates for variance reduction, resulting in faster and more stable convergence, particularly in the presence of data heterogeneity and client sampling. |
| [11] | FedProx | MNIST, FEMNIST, Shakespeare, Sent140 | Average accuracy is more than 22% than FedAvg | The core contribution of this paper is the introduction of FedProx, a novel FL framework that handles statistical and systemic heterogeneity, ensures convergence for non- identical data distributions, and outperforms state-of-the- art FedAvg in robustness and accuracy. |
| [38] | FedDyn | MNIST, EMNIST, CIFAR-10, CIFAR-100 and Shakespeare | Accuracy: MNIST=98.4, EMNIST=95.0, CIFAR-10=85.2, CIFAR-100=55.3, Shakespeare=51.2 | The core contribution of this study is a new FL approach that dynamically regulates the interoperation between a subset of devices, coordinating local and global solutions while minimizing transmission costs and taking into account the heterogeneity of devices. |
| [39] | MOON | CIFAR-10, CIFAR-100, and Tiny-Imagenet | True-Classified Scenario = 90.37, False-Classified Scenario = 9.63 | The core contribution of this paper is the introduction of MOON, a model-contrastive FL approach that leverages model representation similarity to enhance federated training, achieving superior performance on image pattern recognition tasks compared to other state-of-the-art algorithms. |
| [40] | FedGen | MNIST, EMNIST, Celeba | Accuracy: MNIST=95.52, EMNIST=78.92, Celeba=90.29 | The core contribution of this paper is a data-free knowledge distillation approach that enhances heterogeneous FL by utilizing a lightweight generator to ensemble user information, leading to improved generalization performance and reduced communication rounds compared to state-of-the-art methods. |

## 3- Research Methodology

FL algorithms were implemented by first defining the problem intended to be solved through this distributed approach. The problem was outlined, and the participating clients in the FL process were identified. Subsequently, data were collected and preprocessed on each client while ensuring that they remained localized to the device. Once the problem and data were defined, seven PFL algorithms were selected to address the problem effectively.

To facilitate the FL process, we established a central server responsible for coordinating the interactions among clients. Then we developed communication protocols that enable clients to send and receive updates from the server. After that, we initialized the global model on the central server as the starting point for the collaborative training process. We also distributed the initial model weights to all participating clients and guided them to load their local data along with the model weights. The training loop constituted a core part of the workflow. Iterating over multiple rounds, a subset of clients is selected at random in each round. These clients engage in local model training using their own data and the current global model. Gradients are computed during this local training, but the global model is not updated yet. Subsequently, the calculated gradients are sent to the central server by the clients. The central server aggregates these gradients, often using methods like weighted averaging, and updates the global model accordingly.

The global model's performance was monitored and evaluated throughout the training process. The model was periodically assessed on a validation dataset to ensure it improved as expected. In addition, data privacy and security were prioritized by implementing secure aggregation to protect client data during communication and aggregation processes. Finally, hyperparameters were tuned by experimenting with learning rates, aggregation methods, and other factors to enhance training performance and convergence speed. Once the training process concluded, the stopping point was determined based on convergence or a predefined number of rounds. The trained global model was then deployed for PFL assessment tasks. Figure 1 illustrates the overall workflow.

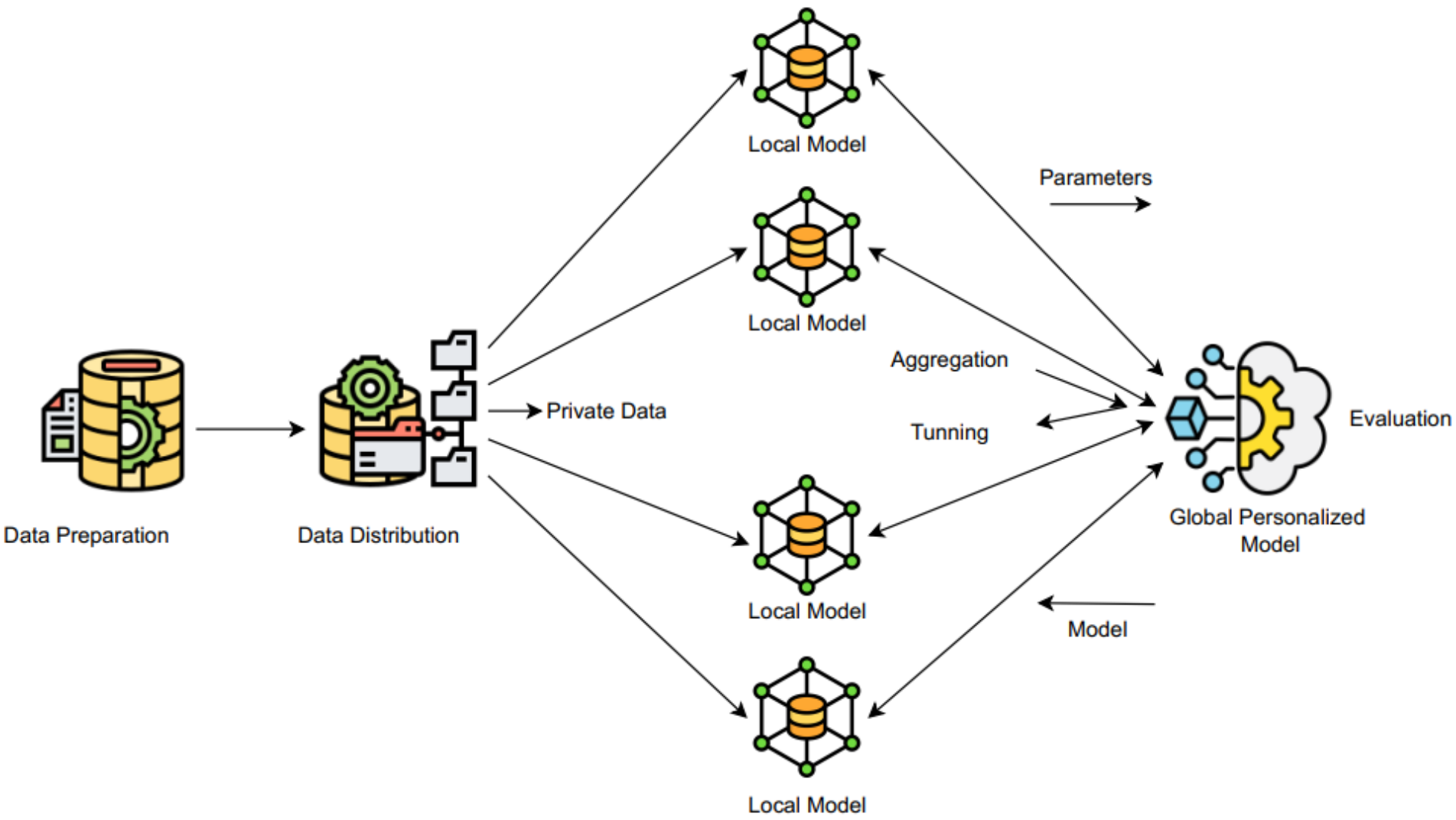

**Figure 1.** **Framework Workflow**

### *3-1- Overview of Datasets*

#### *3-1-1- MNIST*

The MNIST dataset is a widely used benchmark dataset in the field of ML and pattern recognition. It comprises a collection of 70,000 grayscale images of handwritten digits (0 to 9), each having a resolution of 28x28 pixels. The dataset is divided into two main subsets: 60,000 images for training and 10,000 images for testing. Its simplicity and availability have made it an essential resource for developing and evaluating various image recognition algorithms and DL models. Researchers and practitioners often use the MNIST dataset as a starting point to experiment with new techniques due to its manageable size and clear task of digit classification.

#### *3-1-2- Sign-MNIST*

The Sign-MNIST dataset is a valuable benchmark dataset widely used in the fields of sign language recognition and gesture recognition. It consists of a collection of 27,455 grayscale images of hand gestures representing the American Sign Language (ASL) alphabet, including signs for letters A to Z. Each image has a resolution of 28x28 pixels. The dataset is divided into two primary subsets: 25,455 images for training and 2,000 images for testing. Sign-MNIST serves as a fundamental resource for developing and evaluating various ML and DL algorithms for sign language translation and gesture pattern recognition tasks. Its accessibility and well-defined task of sign recognition make it an ideal starting point for researchers and practitioners to explore and innovate in this area of pattern recognition.

#### *3-1-3- Digit-5*

The Digits-5 dataset is a compilation of five popular digit datasets, each containing various styles of 0-9 digit images. The datasets included are MNIST (mt) with 55,000 samples, MNIST-M (mm) with 55,000 samples, Synthetic Digits (syn) with 25,000 samples, SVHN (sv) with 73,257 samples, and USPS (up) with 7,438 samples. These datasets are commonly used in ML and pattern recognition for tasks such as digit recognition and classification. The variety of styles present in the dataset makes it valuable for evaluating and comparing different algorithms and models in the field.

### *3-2- Experimental Setup*

The benchmarks are based on the MNIST [27], Sign-MNIST, and Digit-5 datasets of digit items with non-IID data distributions. For the MNIST dataset, 50 items were selected from a handwritten digit repository, with 5 samples for each of the 10 categories corresponding to the MNIST classes. For the Sign-MNIST dataset, data were obtained from https://www.kaggle.com/datasets/datamunge/sign-language-mnist, and 120 items were selected from the sign language repository, with 5 samples for each of the 24 categories. In addition, the algorithms were tested on the Digit-5 dataset collected from https://drive.google.com/file/d/1A4RJOFj4BJkmliiEL7g9WzNIDUHLxfmm/view.

Four virtual machines connected in a star topology using Ethernet were used for the experiments. Each machine is equipped with an x86_64 architecture running ArchLinux 2023.04.01 with kernel version 6.2.2-arch1-1. Machines A and B utilize an i9-9900K CPU with 31.75 GiB of memory, while machines C and D use an i9-10900K CPU with 31.76 GiB of memory. All machines operate on the Linux operating system. This star topology enables direct connections between the virtual machines, allowing efficient communication.

### 3-3- Data Pre-Processing

Samples assigned to each client were intentionally selected to create non-IID data distributions. This was achieved using the sorting and sharding method [5], which involves sorting the data, dividing it into equal-sized shards, and randomly assigning a specified number of shards to each client. Each client received two shards of size 300 in the datasets, which included 20 clients. Additionally, the 20 clients across all distributed benchmarks were deployed as evenly as possible across the machines, with one machine acting as the server.

### 3-4- Algorithms

#### 3-4-1- Adaptive Personalized Cross-Silo Federated Learning (APPLE)

The algorithm APPLE is a personalized cross-silo FL framework that adaptively learns how much each client can benefit from other clients' models [20]. The APPLE (Adaptive Personalized Cross-Silo Federated Learning) algorithm initializes the clients' base models and parameters; performs personalized local optimization rounds where clients iteratively update the base models and DR vectors using personalized model aggregation, empirical risk calculation, and gradient descent updates; and finally uploads the optimized base models to the central server. This approach enables privacy-preserving, cross-silo learning by collaboratively training personalized models across clients while taking into account empirical risk and proximal conditions. Algorithm 1 presents the working procedural code of the above description.

**Algorithm 1. APPLE**

$N$ clients, learning rates $\eta_1, \eta_2$, number of total rounds $R$, proximal term coefficients $\lambda(r)$, μ, prox-center $p_0$ Personalized models $w_1^{(p)}, w_2^{(p)}, \ldots, w_N^{(p)}$ on the site of the corresponding client.

Initialize core models $w_i^{(c)}$ on server ∀i ∈ $[N]$

Initialize local DR vector $p_i$ ∀i ∈ $[N]$ in parallel r → 1, 2, ..., $R$ $i$ → 1, 2, ..., $N$ **in parallel**

Download core models from server as needed local optimization not converged

Compute personalized model $w_i^{(p)}$:

$$w_i^{(p)} = \sum_{j=1}^{N} p_{i,j} w_j^{(c)}$$

Compute empirical risk $F_i(w_i^{(p)})$:

$$F_i(w_i^{(p)}) = \frac{1}{n_i} \sum_{\xi \in D_i^{tr}} L(w_i^{(p)}; \xi) + \frac{\lambda(r)\mu}{2} ||p_i - p_0||$$

Update $w_i^{(c)}$ and DR vector $p_i$:

$$w_i^{(c)} \rightarrow w_i^{(c)} - \eta_1 \frac{\partial}{\partial w_i^{(c)}} F_i(w_i^{(p)})$$

$$p_i \rightarrow p_i - \eta_2 \frac{\partial}{\partial p_i} F_i(w_i^{(p)})$$

Upload local core model $w_i^{(c)}$ to the server models $w_1^{(p)}, w_2^{(p)}, \ldots, w_N^{(p)}$

#### 3-4-2- Adaptive Local Aggregation for Personalized Federated Learning (FedALA)

FedALA is designed for PFL scenarios [21]. Its goal is to enhance collaboration between N clients while adapting to individual data distributions. The FedALA (Adaptive Local Aggregation) algorithm works through a series of iterations: the server sends the initial global model to the clients, who initialize the local weight matrices. In each iteration, a subset of clients is sampled, and the server sends the previous global model for updating. The clients train the local weights based on the iteration count, possibly iteratively, and apply pruning if necessary. They then compute intermediate local models, refine them further using gradient descent, and forward these updates to the server. The server aggregates the local models received to create a new global model. This process is repeated for T iterations, eventually producing a set of refined local models. FedALA adapts to individual data distributions and converges towards better global models by cooperating in FL scenarios. Algorithm 2 presents the working procedural code of the above description.

**Algorithm 2.** FedALA

N clients, $\rho$: client joining ratio, $L$: loss function, $\Theta_0$: initial global model, $\alpha$: local learning rate, $\eta$: the learning rate in ALA, $s\%$: the percent of local data in ALA, $p$: the range of ALA, and $\sigma(\cdot)$: clip function. Reasonable local models $\widehat{\Theta}_1, \dots, \widehat{\Theta}_N$.

Server initializes and sends $\Theta_0$ to all clients

Clients initialize $W_i^{(p)}$: → 1, ∀i ∈ [N]

t → 1, 2, . . . , $T$ Server samples a subset $I^t$ of clients according to $\rho$ Server sends $\Theta^{t-1}$ to all clients in $I^t$

client i ∈ $I^t$ **in parallel** Client $i$ samples $s\%$ of local data

t = $2W_i^p$ does not converge $W_i^\rho \rightarrow W_i^\rho - \eta \nabla W_i^\rho L(\widehat{\Theta}_i^t, D_i^{s,t}; \Theta^{t-1})$ Clip $W_i^p$ using $\sigma(\cdot)$

$W_i^\rho \rightarrow W_i^\rho - \eta \nabla W_i^\rho L(\widehat{\Theta}_i^t, D_i^{s,t}; \Theta^{t-1})$ Clip $W_i^p$ using $\sigma(\cdot)$

Compute $\widehat{\Theta}_i^t$:

$$\widehat{\Theta}_i^t \rightarrow \widehat{\Theta}_i^{t-1} + \left(\Theta^{t-1} - \widehat{\Theta}_i^{t-1}\right) \odot [1^{|\Theta_0|-p}; W_i^\rho]$$

Update local model $\widehat{\Theta}_i^t$:

$$\Theta_i^t \rightarrow \widehat{\Theta}_i^t + \alpha \nabla_{\widehat{\Theta}_i^t} L\left(\widehat{\Theta}_i^t, D_i; \Theta^{t-1}\right)$$

Server aggregates the global model:

$$\Theta^t \rightarrow \sum_{i \in I^t} \frac{k_i}{\sum_{j \in I^t} k_j} \Theta_i^i$$

Client uploads $\Theta_i^t$ to the server $\widehat{\Theta}_1, \dots, \widehat{\Theta}_N$.

### *3-4-3- Federated Averaging with Body Aggregation and Body Update (FedBABU)*

The algorithm FedBABU is an FL algorithm that only updates the model's body during federated training (the head is randomly initialized but never updated), and the head is adjusted for personalization during the evaluation process [22]. The FedBABU algorithm starts by setting the initial global model parameters, which contain separate components for external tasks and pattern recognition tasks. It continues with a series of FL rounds, each of which involves the selection of subsets of clients based on a fraction criterion. Each round selects a random subset of clients and updates their local models simultaneously. This update involves creating temporary local copies of the global model, applying a function to modify the external task component based on local data, and then aggregating the modified external task components from these clients. Finally, the algorithm merges the updated external task component with the original pattern recognition task component to generate the improved global model. This iterative process facilitates collaborative and efficient federation learning across a distributed network of clients. Algorithm 3 presents the working procedural code of the above description.

**Algorithm 3.** FedBABU

Initial global model $\vartheta_G^0 = \{\vartheta_{G,ext}^0, \vartheta_{G,cls}^0\}, K$: total rounds, $f$: fraction of clients, $N$: total number of clients, $\tau$: number of local epochs, $B$: batch size.

Final global model $\vartheta_G^0 = \{\vartheta_{G,ext}^K, \vartheta_{G,cls}^0\}$.

Initialize $\vartheta_G^0 = \{\vartheta_{G,ext}^0, \vartheta_{G,cls}^0\}$

$k \rightarrow 1,2,\dots,K$ $m \rightarrow \max(\lfloor Nf \rfloor, 1)$

$C^k \rightarrow$ random subset of $m$ clients

Client $C_i^k \in C^k$ **in parallel**

$\vartheta_i^k(0) \rightarrow \vartheta_G^{k-1} = \{\vartheta_{G,ext}^{k-1}, \vartheta_{G,cls}^0\}$ $\vartheta_{i,ext}^k(\tau I_i^k) \rightarrow$ **ClientBodyUpdate**$(\vartheta_i^k(0), \tau)$

Update global model:

$$\vartheta_{G,ext}^k \rightarrow \sum_{i=1}^m \frac{n_{C_i^k}}{n} \vartheta_{i,ext}^k(\tau I_i^k), n = \sum_{i=1}^m n_{C_i^k}$$

$\vartheta_G^K = \{\vartheta_{G,ext}^K, \vartheta_{G,cls}^0\}$

ClientBodyUpdate Function:

end $\vartheta_i^k, \tau I_i^k \rightarrow \left\lceil \frac{n_{C_i^k}}{B} \right\rceil$

local epoch $\rightarrow 1, 2, \dots, \tau$

iteration $\rightarrow 1, 2, \dots, I_i^k$

$\vartheta_{i,ext}^k \rightarrow SGD(\vartheta_{i,ext}^k, \vartheta_{G,cls}^0) \vartheta_{i,ext}^k$

### 3-4-4- Federated Learning with Gradient Correction (FedGC)

FedGC facilitates collaborative model training across multiple clients while addressing the gradient inconsistency problems that can arise in FL [23]. The algorithm uses a server-client architecture and operates in circles.

The FedGC algorithm works on K clients with Sk local data distribution and η learning rates. It uses a circular training approach where the server initializes each client's model with the current global parameters, and the clients perform a local SoftMax computation and then update their models using the global model and local data. The server then aggregates the clients' updates to refine the global model. FedGC handles gradient inconsistency and improves convergence and data protection, while model updates require communication between the server and clients.

Algorithm 4 presents the working procedural code of the above description.

**Algorithm 4. FedGC**

---

$K$ clients indexed by $k$, local data distributed over $S_k$, learning rate $\eta$, regularization parameter $\lambda$.

Final global parameters $\vartheta^T, W^T$.

Server initializes global model parameters $\vartheta_0, W_0$

$t \rightarrow 0, 1, \dots, T\text{-}1$

Server initializes each client model with $\vartheta^T, W_k^t$ client k = 1, 2, …, K **in parallel**

Client k computes local Softmax

Update client model parameters:

$$\vartheta_k^{t+1}, W_k^{t+1} \rightarrow (\vartheta_k^t, W_k^t) - \eta \nabla \ell_k$$

Client sends $(\vartheta_k^{t+1}, W_k^{t+1})$ to the server

Server aggregates global model parameters:

$$\vartheta^{t+1} \rightarrow \sum_{k=1}^{K} \frac{n_k}{n} \vartheta^{t+1}_k, n = \sum_{k=1}^{K} n_k$$

$$\overline{W}^{t+1} \rightarrow \left[W^{t+1}_1, W^{t+1}_2, \dots, W^{t+1}_K\right]^T$$

Server applies gradient correction:

$$W^{t+1} \rightarrow \overline{W}^{t+1} - \lambda \eta \nabla_{\overline{W}^{t+1}} Reg(\overline{W}^{t+1})$$

$\vartheta^T, W^T$

---

### 3-4-5- Personalized Federated Learning with Feature Alignment and Classifier Collaboration (FedPAC)

FedPAC algorithm is designed for PFL with feature alignment and classifier collaboration [24]. It operates over a defined number of communication circles (T) involving multiple clients. The goal is to increase global model performance by feature matching and collaboration on personalized classifiers.

The FedPAC algorithm takes learning rates for feature extraction and classification, communication rounds, client count, local epochs, and a hyper-parameter as inputs. It involves server initialization of global features and centroids, followed by rounds where client sets are selected, global parameters are broadcasted, and client models are collected and updated. The global feature extractor and centroids are recalculated, and personalized classifiers are computed using a solver. Clients then update their local feature extractors, train personalized classifiers and global feature extractors, and return updated parameters to the server.

The algorithm aims for PFL with feature alignment and collaborative classifiers. Algorithm 5 presents the working procedural code of the above description.

**Algorithm 5.** FedPAC

```
Learning rates {η_f, η_g}, number of communication rounds T , number of clients m, number of local epochs E,
hyperparameter λ. Global feature extractor ϑ~T+1, personalized classifiers {ϕ~_i^(T+1)}_(i=1)^m.
```

**Server executes:**

Initialize global parameters: $w^{(0)}, c^{(0)}$

$t \to 0, 1, \ldots, T-1$ **Server:**

Select client set $C_t$ Broadcast $\{w^{(t)}, c^{(t)}\}$ to selected clients Collect models and statistics from clients

Compute global feature extractor $\vartheta^{\sim t+1}$:

$$\vartheta^{\sim t+1} = \frac{1}{\sum_{i=1}^{m} n_{i,k}} \sum_{i=1}^{m} n_{i,k}\ \hat{c}_{i,k}^{(t+1)},\ \forall \mathrm{k} \in [\mathrm{K}]$$

Compute global centroids $c^{t+1}$:

$$c_k^{t+1} = \frac{1}{\sum_{i=1}^{m} n_{i,k}} \sum_{i=1}^{m} n_{i,k}\ \hat{c}_{i,k}^{(t+1)},\ \forall \mathrm{k} \in [\mathrm{K}]$$

Solve for personalized classifier $\phi^{\sim t+1}_{i}$:

$$\alpha_i^* = arg\ min_{\alpha_i} R_i(\alpha_i), s.t. \sum_{i=1}^{m} \alpha_{ij} = 1, \alpha_{ij} \geq 0, \forall j$$

Send $\{\vartheta^{\sim t+1}, \phi^{\sim t+1}_{i}\}$ back to client $i \in C^t$

ClientUpdate Function: end $i, w^{(t)}, c^{(t)}$

Update local feature extractor: $\vartheta_i^{(t)} \to \vartheta_i^{\sim(t)}$

Extract feature statistics $\mu_i^{(t)}, V_i^{(t)}$

Train $\phi_i^{(t+1)}$ for 1 epoch and $\vartheta_i^{(t+1)}$ for E epochs alternately

Extract local feature centroids:

$$\hat{c}_{i,k}^{(t+1)} = \frac{\sum_{l=1}^{n_i} 1\left(y_{(i),l} = k\right) f_{\vartheta_i^{(t+1)}}(x_{(i),l})}{\sum_{l=1}^{n_i}\left(y_{(i),l} = k\right)}, \forall k \in \lfloor K \rfloor$$

$\{\ \vartheta_i^{(t+1)}, \phi_i^{(t+1)}, \hat{c}_i^{(t+1)}, \mu_i^{(t)}, V_i^{(t)}\ \}$ to server

### *3-4-6- Federated Learning from Pre-Trained Models: A Contrastive Learning Approach (FedPCL)*

FedPCL is designed for FL using pre-trained models using a contrastive learning strategy [25]. It exploits pre-trained backbone networks and prototype sets for enhanced performance. FedPCL initially takes local data and model parameters from each client, as well as pre-trained backbone networks. During communication rounds, clients update their local prototype sets. The server then updates the global prototype set. For each client local update, it calculates the losses using the global and local prototypes, updates the model parameters, calculates the local prototypes, and returns the updated prototype set. FedPCL leverages pre-trained models and contrastive learning to enhance FL, which has the benefits of knowledge transfer and feature improvement, but introduces complexity, requires parameter tuning, and comes with communication considerations. Algorithm 6 presents the working procedural code of the above description.

**Algorithm 6.** FedPCL

Client Datasets $D_i$, model parameters $\vartheta_i$ for $i = 1, \ldots, m$, $K$ pre-trained backbones with parameters $\{\phi_1^*, \phi_2^*, \ldots, \phi_k^*\}$, temperature parameter $\tau$.

Updated global and local prototypes $\{\overline{C}^{(j)}\}$ and $\{C_i\}$, and updated client models $\{\vartheta_i\}$.

**Server executes:**

Initialize prototype sets $\{C_p\}_{p=1}^{m}$

$T \to 1, 2, \ldots,$ client $i$ **in parallel** $C_i \to$ LocalUpdate$(i, C, \{C_p\}_{p=1}^{m})$

Update global prototypes:

$$\overline{C}^{(j)} \to \frac{1}{|N_j|} \sum_{i \in N_j} \frac{|D_{i,j}|}{N_j} C_i^{(j)}, \forall j$$

**End server execution.**

LocalUpdate Function:

end $i, C, \{C_p\}_{p=1}^{m}$

local epoch batch in $D_i$

Compute global loss:

$$L_g = \sum_{(x,y) \in D_i} \left( -\log \sum_{y_a \in A(y)} \frac{\exp\left(z_x \cdot \frac{\overline{C}(y)}{\tau}\right)}{\exp\left(z_x \cdot \frac{\overline{C}(y_a)}{\tau}\right)} \right)$$

Compute local loss:

$$L_p = \sum_{(x,y) \in D_i} \left( \sum_{p=1}^{m} \log \sum_{y_a \in A(y)} \frac{\exp\left(z_x \cdot \frac{C_p^{(y)}}{\tau}\right)}{\exp\left(z_x \cdot \frac{C_p^{(y_a)}}{\tau}\right)} \right)$$

Update client model parameters:

$$L(\vartheta_i; z(x), y, C, \{C_p\}_{p=1}^{m} = L_g + L_p$$

Compute local prototypes:

$$C_i^j \to \frac{1}{|D_{i,j}|} \sum_{(x,y) \in D_{i,j}} z(x), \forall j$$

$C_i$

### *3-4-7- Federated Prototype (FedProto)*

The algorithm, FedProto, is a federated prototype learning framework in which the clients and server communicate abstract class prototypes rather than gradients [26]. FedProto aggregates the local prototypes gathered from various clients, then sends the global prototypes back to all clients to regularize the training of local models. While keeping the resulting local prototypes sufficiently close to the corresponding global ones, the training on each client aims to minimize the pattern recognition error on the local data. FedProto first initializes the local data and model parameters for each client, as well as the global prototype set for all classes. During communication rounds, each client updates its local prototype set using a local update function. The global prototype set is then updated by averaging the local prototypes of each class. The local prototype set of each client is then updated with the prototypes of the global set. The algorithm aims to improve federation learning using prototype-based methods, focusing on global-local fusion to improve model performance. Algorithm 7 presents the working procedural code of the above description.

**Algorithm 7. FedProto**

Client datasets $D_i$, model parameters $\omega_i$ for $i = 1, \ldots, m$.

Global prototype set $\{\bar{C}^{(j)}\}$, Updated local prototypes $\{Ci\}$, and model parameters $\{\omega_i\}$.

**Server executes:**

Initialize global prototype set $\{\bar{C}^{(j)}\}$ for all classes

$T \rightarrow 1, 2, \ldots$ client i **in parallel** $C_i \rightarrow$ LocalUpdate$(i, \bar{C}_i)$

Update global prototypes:

$$\bar{C}^{(j)} \rightarrow \frac{1}{|N_j|} \sum_{i \in N_j} \frac{|D_{i,j}|}{N_j} C_i^{(j)}, \ \forall j$$

Update local prototype sets $C_i$ with global prototypes $\{\bar{C}^{(j)}\}$ **End server execution.**

LocalUpdate Function:

end $i, \bar{C}_i$

local epoch batch $(x_i, y_i) \in D_i$

Compute local prototype:

$$C_i^{(j)} \rightarrow \frac{1}{|D_{i,j}|} \sum_{(x,y) \in D_{i,j}} f_i(\phi_i; x), \ \forall j$$

Compute Loss:

$$L(D_i, \omega_i) = L_S(F_i(\omega_i; x), y) + \lambda . L_R(\bar{C}_i, C_i)$$

Update local model parameters $\omega_i$ based on the computed loss $C_i$

## 4- Results

### *4-1- Analysis*

Table 3 provides a comparative analysis of the performance of different FL algorithms on three distinct datasets: MNIST, Sign MNIST, and Digit5. These algorithms include APPLE, FedALA, FedBABU, FedGC, FedPAC, FedPCL, and FedProto. The evaluation metrics used to assess their performance are Accuracy, Precision, Recall, and F1 Score.

APPLE stands out as the top-performing algorithm, consistently achieving outstanding results across all datasets. Its accuracy, precision, recall, and F1 score remain remarkably high, surpassing 99% on MNIST and Digit5 datasets, whereas a comparatively lower results of over 92% in SignMNIST dataset for all evaluation metrics. This performance showcases the algorithm's robustness and effectiveness in a range of tasks.

FedALA exhibits impressive results on the Sign MNIST dataset, achieving an accuracy of 99.88% and an F1 score of 99.89%. However, its performance on MNIST and Digit5 datasets is comparatively lackluster, with accuracy hovering around 41% to 55%. This discrepancy suggests that FedALA might be better suited for specific tasks, such as sign language digit recognition.

FedGC proves to be a reliable choice with consistent high performance across all datasets. Its accuracy consistently exceeds 97%, and the F1 score remains around 98%. This indicates that FedGC possesses strong generalization capabilities and maintains stable performance across various scenarios.

FedPAC displays competitive performance, achieving accuracy ranging from 86% to 91% across all datasets. This consistency suggests that FedPAC is a well-rounded algorithm suitable for diverse FL applications.

In contrast, FedPCL's performance fluctuates considerably. While it demonstrates promising results on MNIST and Sign MNIST datasets, with accuracy above 82%, its performance on the Digit5 dataset drops significantly, achieving an accuracy of only 18.12%.

FedProto proves to be another strong contender, with an accuracy above 94% and an F1 score around 97% on all datasets. This indicates its effectiveness in handling various tasks in the context of FL.

Although the choice of algorithm should be made judiciously, considering the specific dataset and task requirements to optimize performance effectively, algorithms like APPLE, FedGC, and FedProto emerge as the most reliable performers across the evaluated datasets, while others may excel in specific contexts or require further refinement to achieve optimal results.

Figures 2 to 5 in brief, revealed comparable functionality of Table 3 with different parameters on different datasets. Here, Table 4 shows the advantages and disadvantages of each of the PFL algorithms we used for the performance analysis.

**Table 3. Evaluation Matrices Performance of Different FL Algorithms**

| Algorithm | Dataset | Accuracy (%) | Precision (%) | Recall (%) | F1 Score (%) |
|---|---|---|---|---|---|
| APPLE | MNIST | 99.77 | 99.80 | 99.77 | 99.78 |
| | Sign MNIST | 92.68 | 92.71 | 92.68 | 92.79 |
| | Digit5 | 99.66 | 99.69 | 99.66 | 99.67 |
| FedALA | MNIST | 41.94 | 41.97 | 41.94 | 41.95 |
| | Sign MNIST | 99.88 | 99.91 | 99.88 | 99.89 |
| | Digit5 | 92.44 | 92.47 | 92.44 | 92.45 |
| FedBABU | MNIST | 55.79 | 55.82 | 55.79 | 55.80 |
| | Sign MNIST | 44.49 | 44.52 | 44.49 | 44.50 |
| | Digit5 | 49.67 | 49.70 | 49.67 | 49.68 |
| FedGC | MNIST | 97.07 | 97.10 | 97.07 | 97.08 |
| | Sign MNIST | 97.98 | 97.99 | 97.98 | 97.99 |
| | Digit5 | 98.13 | 98.17 | 98.13 | 98.14 |
| FedPAC | MNIST | 86.46 | 86.49 | 86.46 | 86.47 |
| | Sign MNIST | 89.88 | 89.91 | 89.88 | 89.89 |
| | Digit5 | 91.83 | 91.86 | 91.83 | 91.84 |
| FedPCL | MNIST | 82.75 | 82.78 | 82.75 | 82.76 |
| | Sign MNIST | 84.65 | 84.68 | 84.65 | 84.66 |
| | Digit5 | 18.12 | 18.15 | 18.15 | 18.13 |
| FedProto | MNIST | 97.13 | 97.16 | 97.13 | 97.14 |
| | Sign MNIST | 96.70 | 96.73 | 96.70 | 96.71 |
| | Digit5 | 94.93 | 94.96 | 94.93 | 94.94 |

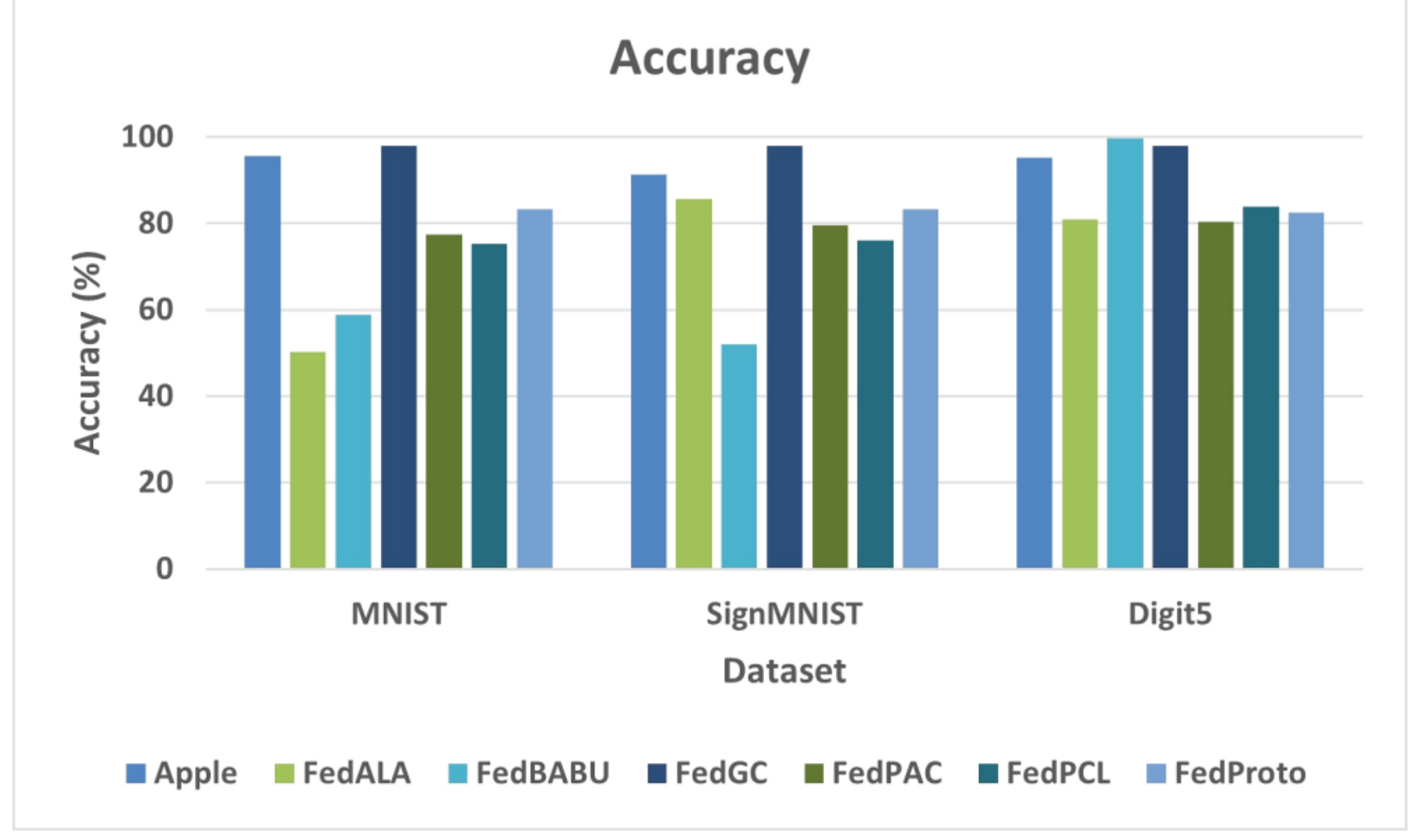

**Figure 2. Accuracy graph of different FL algorithms in MNIST, SignMNIST and Digit5 dataset**

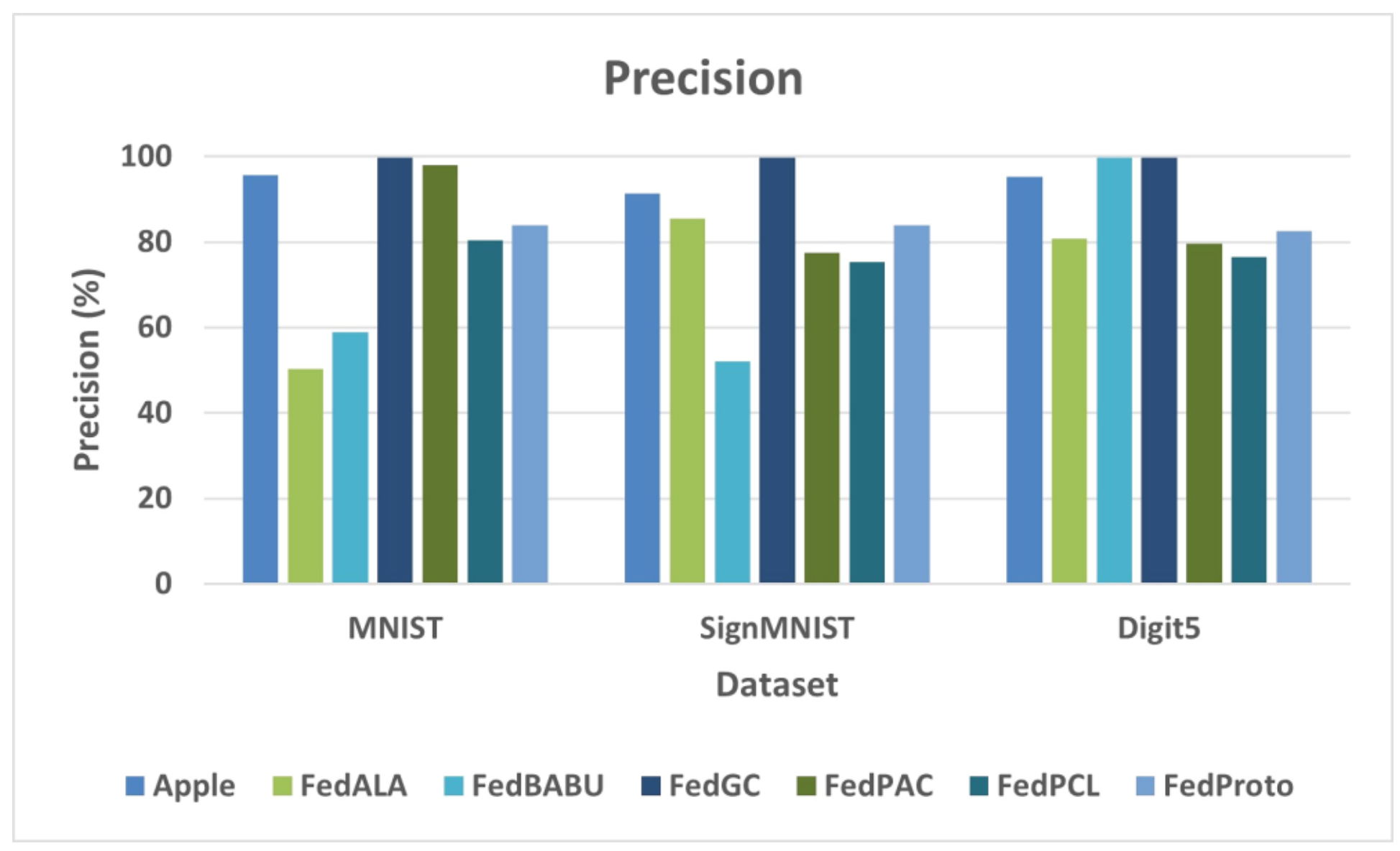

**Figure 3.** Precision graph of different FL algorithms in MNIST, SignMNIST and Digit5 dataset

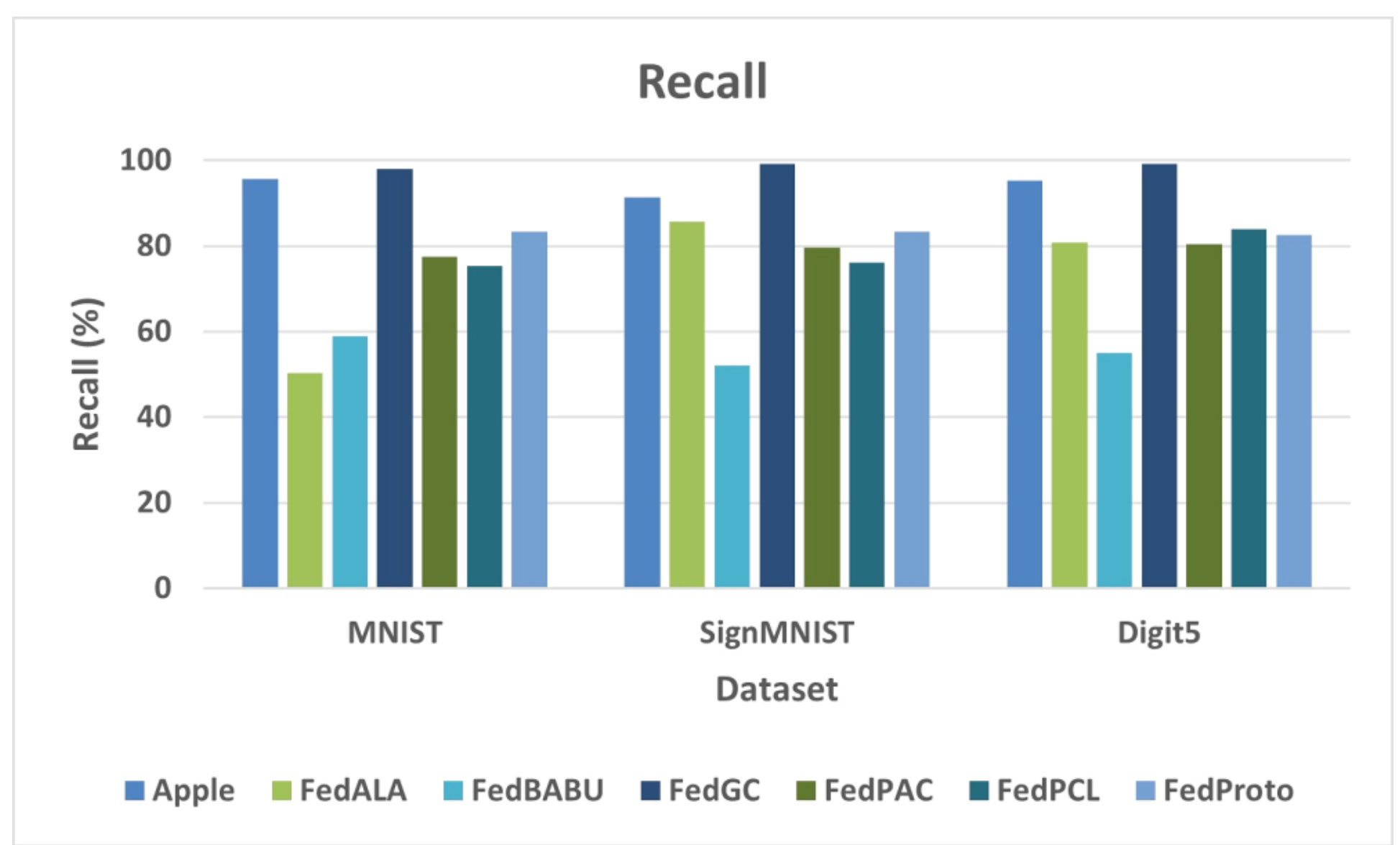

**Figure 4.** Recall graph of different FL algorithms in MNIST, SignMNIST and Digit5 dataset

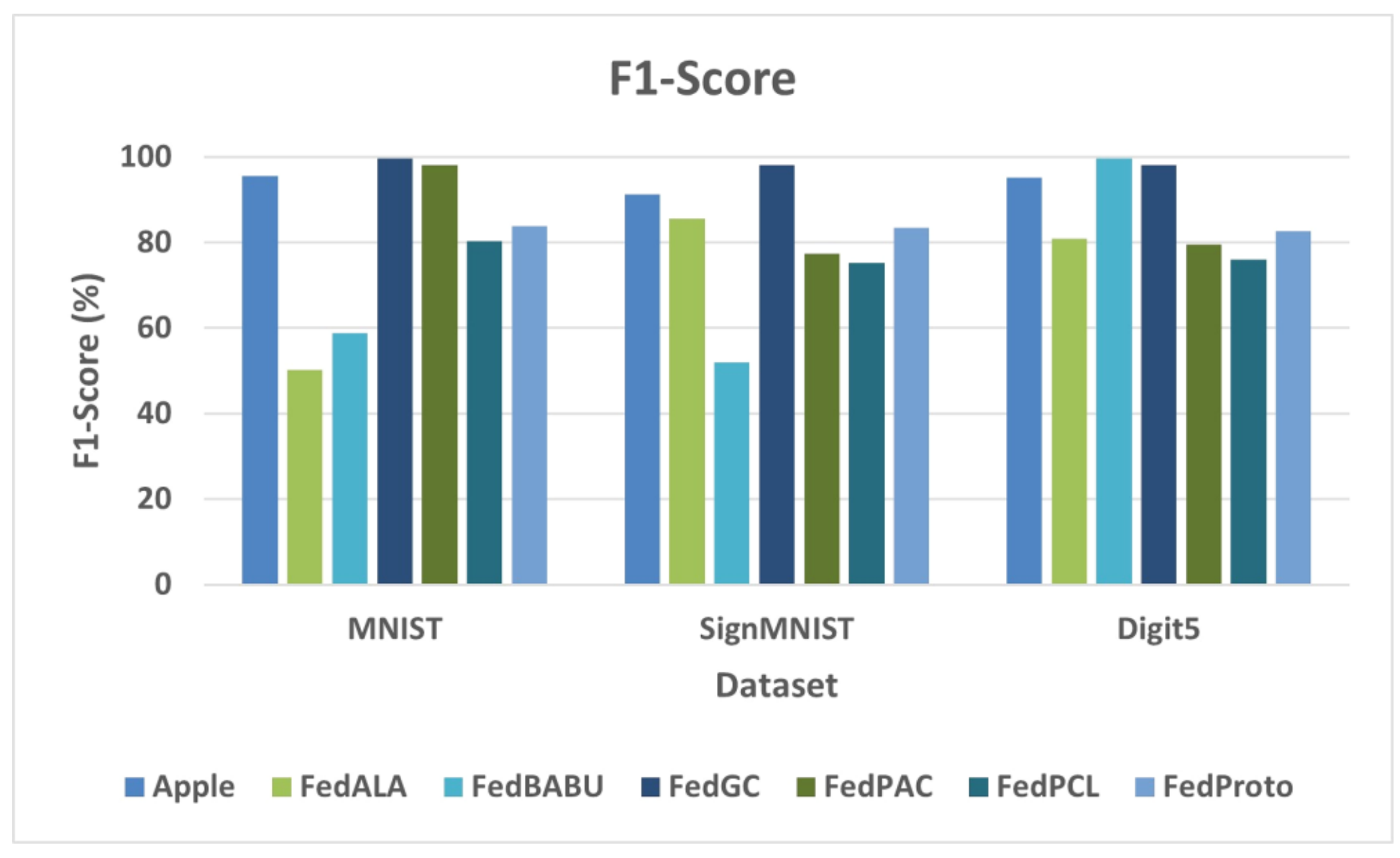

**Figure 5.** F1 Score graph of different FL algorithms in MNIST, SignMNIST and Digit5 dataset

**Table 4. Analysis of advantages and disadvantages of PFL algorithms**

| Algorithm | Dataset | Accuracy (%) |
|---|---|---|
| APPLE | APPLE creates personalized models for each customer, enhancing adaptability to individual data distributions while ensuring data privacy and promoting cross-silo collaboration. The algorithm's proximal term aids stable convergence towards a central point during optimization. | APPLE approach includes slower convergence due to decentralized learning, hyperparameter sensitivity impacting convergence speed and performance, communication overhead from frequent updates, and difficulties in aggregating models with heterogeneous data distributions. |
| FedALA | FedALA introduces PFL with adaptive local aggregation (ALA), dynamically adjusting aggregation techniques based on convergence progress. Integrated with client sampling, local learning rates, and clip functions, it enables efficient personalization, faster convergence, and robustness across heterogeneous clients. | FedALA's use of client sampling and dynamic aggregation may introduce challenges in parameter management, potentially causing sensitivity or computational overhead. While the ALA strategy enhances customization and convergence, its effectiveness relies on well-tuned hyperparameters and local data sampling extent, posing practical implementation and performance complexities. |
| FedBABU | FedBABU innovates FL via body-merging and body- updating, improving communication efficiency and convergence speed. Selective client updates reduce communication overhead for effective model aggregation. Hybrid local and global updates enhance client collaboration, promising better large-scale performance. | FedBABU's reliance on random client selection and subset updates could result in uneven model representation and slower convergence in certain rounds, potentially affecting adaptability to changing data or network conditions. Additionally, the integration of body merging and updating techniques may introduce complexity, necessitating meticulous parameter tuning for optimal performance. |
| FedGC | FedGC improves convergence and model performance by addressing gradient inconsistency through local calculations and global model updates, while ensuring data security through FL. | Potential drawbacks, however, include the additional communication cost due to frequent model changes, the complexity of the algorithm, which requires careful implementation and parameter tuning, and sensitivity to the consistency of data distribution across clients, which can potentially impact the quality of the aggregate model. |
| FedPAC | FedPAC offers personalized learning, feature alignment and training of personalized classifiers to improve the performance of individual models. It exploits feature alignment techniques, promoting coherence between global and local features, and supports collaborative classifier training, facilitating knowledge sharing between clients for better convergence. | The intricacies involved in the algorithm, encompassing both feature alignment and the computation of solver-based classifiers, may give rise to implementation hurdles and the need for nuanced parameter tuning. Communication costs between the server and clients, especially in numerous or slower network scenarios, create difficulties and the quality of the solver affects the effectiveness of the personalized classifier, potentially impacting convergence and performance. |
| FedPCL | FedPCL uses a pre-trained backbone network for efficient knowledge transfer and initialization in federal learning. Contrastive learning improves model performance by encouraging feature proximity for similar instances and separation for different instances. The algorithm allows local adaptation for client- specific fine-tuning within collective learning. | The complexity of the algorithm increases with contrastive learning and pre-training models, requiring efficient training strategies. Contrastive loss, learning rate and hyperparameter adjustment of prototype updates affect convergence and performance. The communication requirements of FL impose additional costs, especially for many clients or slower networks. |
| FedProto | FedProto improves federation learning with prototype-based methods, with an emphasis on class-specific updates for better generalization and robustness. It balances global and local prototypes to achieve synergy between collective knowledge and customer-specific insights. | The computational complexity of the algorithm, including prototype upgrades and additional communication costs, can have an impact on the efficiency of training. Hyperparameters associated with prototype updates and averaging also affect convergence and performance. |

### *4-2- Comparisons of the Analysis with Traditional Machine Learning and Federated Learning Algorithms*

Tables 1 and 2 present the performance of different traditional machine learning (ML) algorithms and federated learning (FL) algorithms, respectively, across various pattern recognition tasks. The reported accuracy values in both tables provide insight into the effectiveness of these algorithms on their respective datasets. Table 3 further presents a set of accuracy values for different FL algorithms across multiple datasets, offering a more comprehensive understanding of their performance.

A comparison of accuracy values across the three tables reveals several notable trends. In Table 1, traditional ML algorithms such as DP-SVM, CNN, KNN, Gaussian Naïve Bayes, Decision Tree, GPFL, and Hybrid-RF exhibit a wide range of performance levels, with accuracy values ranging from the mid-40% to the high 90%, reflecting their varying capabilities across different tasks. In contrast, Table 2 focuses on FL algorithms, including FedAvg, SCAFFOLD, FedProx, FedDyn, MOON, and FedGen, which generally achieve higher accuracy levels than those reported in Table 1, particularly for more complex datasets such as CIFAR-10 and CIFAR-100.

Table 3 provides a detailed comparison of accuracy performance for various personalized federated learning (PFL) algorithms. The reported accuracy values are competitive with those of traditional ML algorithms in Table 1 and, in some cases, surpass them. For example, algorithms such as APPLE and FedALA achieve accuracy levels in the high 90% range and even approach 100% on datasets such as MNIST, Sign-MNIST, and Digit-5. These results indicate that certain FL algorithms can match or exceed the performance of traditional ML methods. Overall, the comparison across the tables highlights a promising trend in the accuracy performance of FL algorithms. While both traditional ML and FL approaches demonstrate strong capabilities in pattern recognition tasks, PFL algorithms show significant potential to achieve competitive—and in some cases superior—accuracy, particularly in distributed data environments with heterogeneous data distributions.

The above comparative analysis of PFL algorithms provides a detailed view of the performance across various datasets, with a focus on accuracy, precision, recall, and F1 score. However, while the analysis emphasizes the competitive performance of these algorithms, it does not fully highlight the significant trade-offs between accuracy, customization, and data security, which are central to the design and deployment of FL systems.

## 5- Trade-Offs Between Accuracy, Customization, and Data Security

FL algorithms, particularly in the personalized setting, typically strive to balance three core aspects: accuracy, customization, and data security. The trade-offs between these aspects are not always straightforward, and the choice of algorithm depends on the specific needs and constraints of the task at hand.

### *5-1- Accuracy vs. Customization*

Many of the algorithms, such as APPLE and FedProto, demonstrate impressive accuracy by combining local adaptation and global model aggregation. However, these gains in accuracy often come at the expense of customization for individual clients. For instance, while APPLE achieves near-perfect accuracy on datasets like MNIST and Digit5, its approach, which balances centralized and decentralized updates, may not fully capture the nuances of personalized data distributions across clients. Similarly, algorithms like FedPAC that aim for better customization through personalized classifiers may experience slightly lower accuracy compared to algorithms that focus more on global aggregation like FedGC, as fine-tuning individual models can introduce inefficiencies.

This trade-off is especially pronounced when the data distributions across clients vary significantly. Algorithms like FedBABU or FedProto that focus on shared representations or prototype-based updates may be less effective in handling highly heterogeneous data from different clients, leading to a loss in model performance despite their robustness.

### *5-2- Customization vs. Data Security*

The core advantage of FL is its ability to preserve data security by ensuring that data does not leave local devices. However, enhancing customization like via personalized models or techniques like FedPAC or FedPCL can sometimes create vulnerabilities. The need to frequently communicate personalized model updates between the clients and the central server can increase the risk of data leakage or adversarial attacks. While techniques like secure aggregation or differential privacy can mitigate some of these risks, the additional computational and communication overheads may reduce the algorithm's efficiency.

### *5-3- Accuracy vs. Data Security*

A significant challenge in FL is maintaining high accuracy while ensuring robust data security. For example, FedGC performs well on global datasets but may not offer the same level of security for highly sensitive data as algorithms designed to focus on privacy, such as those using encryption methods. The trade-off between these two becomes especially apparent in healthcare or financial applications, where high accuracy is necessary but the risk of revealing sensitive information through model updates must be minimized.

The more personalized an algorithm is, the more likely it is to trade off global accuracy for local customization. While personalized models, like those from FedPAC or FedProto, can perform better for individual clients, they may not scale as well when the system needs to aggregate insights from a diverse range of clients. On the other hand, algorithms that focus on global aggregation, such as FedGC, may achieve higher overall accuracy but could fail to address individual client needs, leading to less personalized experiences. The challenge lies in finding the right balance for the specific context of deployment, where the advantages of personalization can be leveraged without compromising global performance or security.

## 6- Conclusion

The results of this study provide a comprehensive performance comparison of various PFL algorithms against traditional ML and FL methods in pattern recognition tasks. The analysis of multiple algorithms, including APPLE, FedGC, and FedProto, across tested datasets reveals that PFL algorithms significantly outperform traditional ML and general FL approaches, especially in terms of accuracy, precision, recall, and F1 score. APPLE, in particular, emerges as the top performer, demonstrating exceptional reliability across diverse datasets, with accuracy exceeding 99% on MNIST and Digit5. FedGC and FedProto also consistently exhibit high performance, with accuracy scores above 97%. In contrast, some FL algorithms, such as FedBABU and FedPCL, show more variable or lower performance, highlighting the necessity of choosing the right algorithm based on specific data and task requirements. FedALA stands out on Sign MNIST, suggesting that certain algorithms may excel in specialized contexts. Moreover, the study underscores the importance of model selection in FL, with PFL algorithms providing enhanced privacy, security, and adaptability to heterogeneous data distributions compared to traditional ML methods.

## 7- Declarations

### *7-1- Author Contributions*

Conceptualization, A.A.N. and M.J.H.; methodology, M.R.A.; software, A.A.; validation, M.A.R., I.D., and M.R.A.; formal analysis, M.J.H.; investigation, M.R.A.; resources, B.M.T.H.; data curation, A.A.; writing—original draft preparation, M.R.A.; writing—review and editing, M.A.R., I.D., M.R.A., B.M.T.H., A.A.N., A.A., and M.J.H.; visualization, I.D.; supervision, B.M.T.H.; project administration, M.A.R.; funding acquisition, M.J.H. All authors have read and agreed to the published version of the manuscript.

### *7-2- Data Availability Statement*

The data presented in this study are available in the article.

### *7-3- Funding*

This research received financial support from Multimedia University, Melaka 75450, Malaysia.

### *7-4- Institutional Review Board Statement*

Not applicable.

### *7-5- Informed Consent Statement*

Not applicable.

### *7-6- Conflicts of Interest*

The authors declare that there is no conflict of interest regarding the publication of this manuscript. In addition, the ethical issues, including plagiarism, informed consent, misconduct, data fabrication and/or falsification, double publication and/or submission, and redundancies have been completely observed by the authors.

## 8- References